\documentclass[10pt,twocolumn,letterpaper]{article}

\usepackage{cvpr}
\usepackage{svg}
\usepackage{multirow}
\usepackage{utfsym}
\usepackage{bbding}
\usepackage[normalem]{ulem}
\useunder{\uline}{\ul}{}
\usepackage{algorithm}
\usepackage{listings}
\definecolor{cvprblue}{rgb}{0.21,0.49,0.74}
\usepackage[pagebackref,breaklinks,colorlinks,allcolors=cvprblue]{hyperref}

\def\paperID{} % *** Enter the Paper ID here
\def\confName{CVPR}
\def\confYear{2026}

\title{Mono3DV: Monocular 3D Object Detection with 3D-Aware Bipartite Matching and Variational Query DeNoising}

\author{
    Kiet Dang Vu$^1$ \qquad
    Trung Thai Tran$^1$ \qquad
    Kien Nguyen Do Trung$^1$ \qquad
    Duc Dung Nguyen$^1$\thanks{Corresponding author.} \\
    $^1$Ho Chi Minh University of Technology, VNUHCM \\
    {\tt\small \{kiet.dangvutuan0712, thai.tran241002, kien.nguyen2211724, nddung\}@hcmut.edu.vn}
}

\begin{document}
\maketitle
\begin{abstract}
\label{sec:abstract}
    % DETR-like architectures have shown great promise in monocular 3D object detection. However, a key challenge is that existing methods often discard the 3D attributes during the Bipartite Matching process. This omission stems from the inherent instability of depth estimation in monocular images. Consequently, high-quality 3D predictions can be incorrectly discarded from the training loss by the resultant 2D-only matching, leading to suboptimal results. To address this, we present Mono3DV, a novel Transformer-based monocular 3D object detection method. First, we introduce the 3D-Aware Bipartite Matching, which effectively incorporates 3D information into the matching process, directly resolving the mismatch inherent in naive 2D-only approaches. Furthermore, we introduce 3D DeNoising to stabilize the Bipartite Matching, thereby mitigating the instability that occurs when integrating 3D attributes caused by ill-posed monocular depth estimation. Recognizing the gradient vanishing issue associated with conventional denoising techniques, we propose a novel Variational Query DeNoising mechanism to overcome this limitation, which significantly enhances model performance. Our method achieves state-of-the-art performance on the KITTI benchmark without requiring any external training data.
    While DETR-like architectures have demonstrated significant potential for monocular 3D object detection, they are often hindered by a critical limitation: the exclusion of 3D attributes from the bipartite matching process. This exclusion arises from the inherent ill-posed nature of 3D estimation from monocular image, which introduces instability during training. Consequently, high-quality 3D predictions can be erroneously suppressed by 2D-only matching criteria, leading to suboptimal results. To address this, we propose Mono3DV, a novel Transformer-based framework. Our approach introduces three key innovations. First, we develop a 3D-Aware Bipartite Matching strategy that directly incorporates 3D geometric information into the matching cost, resolving the misalignment caused by purely 2D criteria. Second, it is important to stabilize the Bipartite Matching to resolve the instability occurring when integrating 3D attributes. Therefore, we propose 3D-DeNoising scheme in the training phase.
    Finally, recognizing the gradient vanishing issue associated with conventional denoising techniques, we propose a novel Variational Query DeNoising mechanism to overcome this limitation, which significantly enhances model performance. Without leveraging any external data, our method achieves state-of-the-art results on the KITTI 3D object detection benchmark.

\end{abstract}
\begin{figure*}[ht!]
    \centering
    \includegraphics[width=\textwidth]{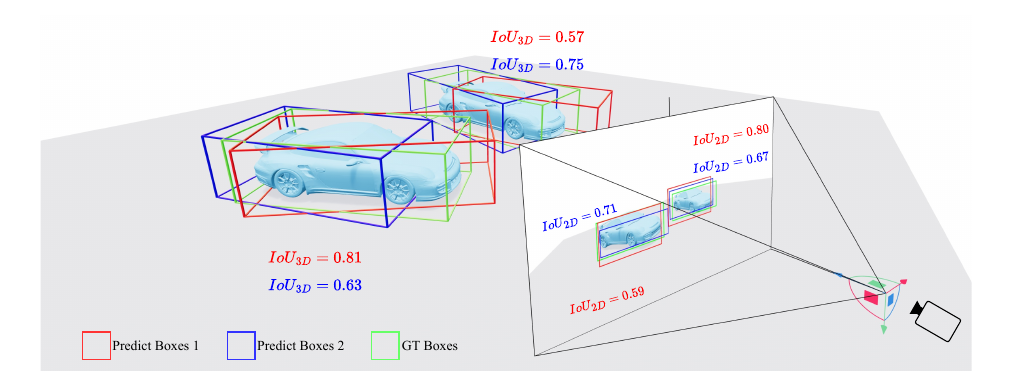}
    \caption{\textbf{Limitations of 2D-only Bipartite Matching.} A major limitation of using a 2D-only matching cost for bipartite assignment is that it prioritizes high-quality 2D predictions even if their associated 3D bounding boxes are poor. Conversely, superior 3D predictions are often discarded if their 2D projection is merely sufficient. This leads to suboptimal optimization because the model is trained based on 2D fidelity rather than the desired 3D accuracy.}
    \label{fig:motivation_3d_matching}
\end{figure*}

\section{Introduction}
\label{sec:intro}
Accurate 3D object detection stands as a cornerstone of autonomous driving systems, providing the essential capability to precisely perceive and understand the surrounding environment. This understanding, encompassing the precise localization, dimensional attributes, and spatial orientation of crucial objects such as vehicles and pedestrians, is paramount for ensuring safe navigation and enabling well-informed decision-making processes within autonomous vehicles. While methodologies leveraging the high-fidelity depth information offered by LiDAR sensors~\cite{PeP,Centerbased3D,PV-RCNN} and sophisticated multi-camera configurations~\cite{RayDN,BEVFormer,PETR} have showcased superior performance in this domain, these multi-sensor approaches inherently present certain limitations. Their dependence on multiple sensing modalities introduces increased system complexity, elevates the potential for sensor failures or miscalibration issues, and consequently can restrict their widespread deployment, particularly in cost-sensitive application scenarios. Therefore, monocular 3D object detection emerges as a highly compelling, inherently robust, and practically advantageous alternative for resource-constrained deployments, as it requires the utilization of only a single camera sensor.

Despite advancements in monocular 3D object detection~\cite{DCD,DID-M3D,MonoCD,MonoUNI,MonoJSG,MonoDDE}, the inherent lack of direct depth information from single-view images remains a significant challenge. To mitigate this limitation, several studies have focused on incorporating estimated depth maps to guide the detection learning process~\cite{MonoDTR,MonoDETR,MonoPGC,FD3D}. Notably, MonoDETR~\cite{MonoDETR} pioneered a DETR-based framework for monocular 3D detection, using a depth-guided transformer. Although this architecture significantly improved object localization over prior methods, its reliance on a Bipartite Matching setup with only 2D attributes creates an inherently suboptimal optimization. Specifically, as illustrated in \cref{fig:motivation_3d_matching}, the matching process can prioritize a candidate prediction with a superior 2D bounding box, even if its associated 3D bounding box is inferior. This mismatch causes the better 3D prediction to be incorrectly discarded from the training loss by the 2D-only matching mechanism.

Although MonoDETR~\cite{MonoDETR} initially integrated the 3D attribute into the Bipartite Matching, naively combining the 3D attribute can lead to training collapse, which is caused by instability in 3D prediction in the early training stage. To address these issues, we proposed 3D-Aware Bipartite Matching, which incorporates the 3D attribute into the Bipartite Matching process via a scheduler. Additionally, we introduced 3D DeNoising to mitigate the instability that arises when using the 3D attribute for Bipartite Matching due to the ill-posed nature of 3D estimation from monocular images. Furthermore, we observed the gradient vanishing problem encountered with conventional denoising approaches and presented Variational Query DeNoising to overcome it, which significantly enhances the model's performance.

In summary, we propose a transformer-based method
called Mono3DV. Our contributions are listed as follows:
\begin{itemize}
    \item We introduce a novel matching mechanism, 3D-Aware Bipartite Matching, that effectively incorporates the 3D attribute using a scheduler, thereby solving the mismatch problem inherent in naive 2D-only approaches.
    \item We introduce 3D-DeNoising to stabilize the Bipartite Matching process, correcting the instability that occurs when integrating 3D attributes compromised by the ill-posed nature of 3D estimation from monocular images.
    \item We observed the gradient vanishing issue associated with conventional denoising and presented a  Variational Query DeNoising to overcome it, which significantly enhances model performance.
    \item Evaluated on the KITTI 3D object detection benchmark, without any extra data, Mono3DV achieves the state-of-the-art performance among monocular detectors.
\end{itemize}

\section{Related Work}
\label{sec:related_work}
\textbf{Monocular 3D Object Detection.} Inferring object depth from single 2D images is a challenge for monocular 3D detection. Researchers explore diverse methods to mitigate this. Initial approaches~\cite{monoflex,DCD,MonoCD} enhanced depth by generating multiple candidates from geometry-inspired techniques and applying depth ensembling for a refined value. More recent efforts~\cite{MonoDTR,MonoDETR,MonoPGC,FD3D} integrate transformer architectures. These methods typically extract visual and depth features using a backbone and lightweight predictor, which are then processed by transformer encoders and aggregated in the decoder for robust detection. MonoDETR~\cite{MonoDETR} represents an initial application of the DETR framework to monocular 3D object detection. It achieves this by predicting foreground depth, integrating a depth-guided decoder, and employing object queries for effective global feature aggregation. However, a key limitation stems from its naive 2D-only Bipartite Matching scheme, which can erroneously suppress high-quality 3D predictions, ultimately leading to suboptimal performance. In this research, we introduce 3D-Aware Bipartite Matching to directly address and resolve the inherent mismatch problem encountered in conventional 2D-only association approaches.

\noindent\textbf{Detection Transformer.} Since the introduction of the Detection Transformer (DETR)~\cite{DETR}, significant progress has been made in object detection. Subsequent work has addressed its limitations, such as slow convergence and limited spatial resolution, through various innovations. Deformable DETR~\cite{DeformableDETR} replaced the original attention with deformable attention for more efficient feature sampling. DN-DETR~\cite{DN-DETR} introduced a denoising training scheme to stabilize the Bipartite matching from inconsistent optimization goals. GroupDETR~\cite{GroupDETR} further improved training stability and performance by incorporating one-to-many matching methods, providing additional positive supervision. In this work, we extend these advancements to monocular 3D object detection, with a particular focus on enhancing the denoising strategy to resolve the instability that occurs when integrating 3D attributes compromised by the ill-posed nature of 3D estimation from monocular images.

\noindent\textbf{Multi-task learning.} Multi-task learning is a widely studied topic in computer vision. Many works focus on adjusting weights for different loss functions to solve the multi-task problem \cite{GradNorm,MTLU}. GradNorm \cite{GradNorm} aimed to resolve the loss unbalance problem in joint multi-task learning, leading to improved training stability. Kendall \etal~\cite{MTLU} proposed a task-uncertainty strategy to address task balance issues, which also achieved strong results. Notably, for monocular 3D object detection, GUPNet \cite{GUPNet} introduced a Hierarchical Task Learning strategy based on task dependencies, ensuring each task begins training only after its designated prerequisite task has been sufficiently optimized. To address the inherent mismatch problem in naive 2D-only Bipartite Matching, this work introduces 3D-Aware Bipartite Matching that balances 3D and 2D costs. Recognizing that the 3D cost is often high and unstable during the early training phase, we propose a novel scheduler that gradually increases the weight of the 3D cost. This strategy ensures that the 3D estimation is only incorporated into the matching process as its accuracy improves throughout training.
\section{Method}
\label{sec:method}
\begin{figure*}[t!]
    \centering
    \includegraphics[width=1.1\textwidth]{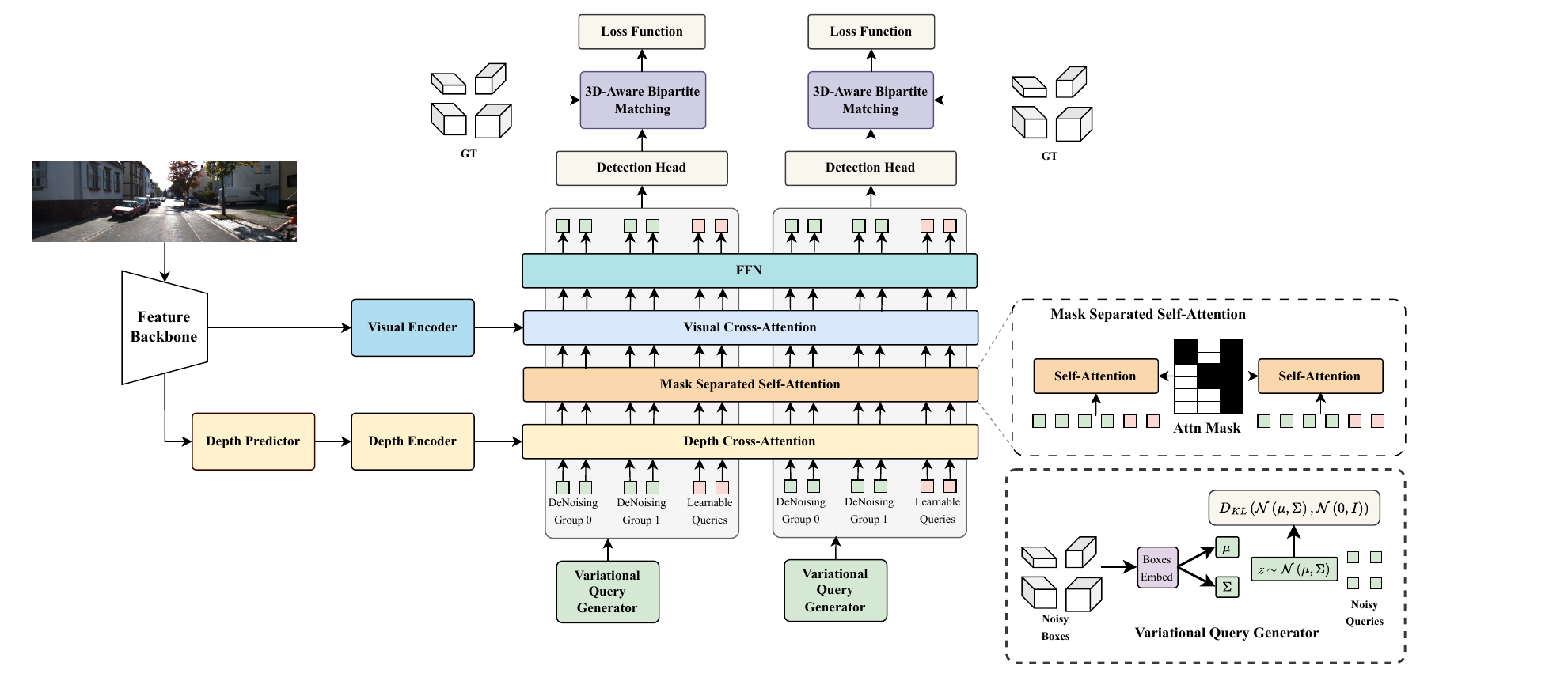}
    \caption{\textbf{The overall of our proposed framework Mono3DV.} The architecture initially extracts features from a single-view image using an image backbone, which are then fed into a Transformer encoder. The subsequent decoder utilizes both standard learnable queries and supplementary noisy queries generated by a Variational Query Generator. Finally, the loss is determined by subjecting the predictions derived from the learnable queries to 3D-Aware Bipartite Matching.}
    \label{fig:overall}
\end{figure*}
\subsection{Overview}
\label{sec:method:overview}
\cref{fig:overall} illustrates the architecture of our proposed framework. Given a single-view image, we first employ ResNet50~\cite{ResNet} to extract high-level feature maps, which are subsequently input to a transformer encoder. Following the baseline MonoDETR\cite{MonoDETR}, the decoder utilizes $G$ groups of learnable queries $Q_L = \{q_{L_i}\}_{i=1}^G$, where $q_{L_i}\in\mathbb{R}^{N \times D}$, with $N$ representing the number of queries in each group and $D$ denoting the hidden dim. For each of these $G$ groups of learnable queries, we introduce $C$ corresponding groups of noisy queries $Q_N = \{\{q_{N_{ij}}\}_{j=1}^C\}_{i=1}^G$, where for the $i$-th group of learnable queries, we have $C$ associated groups of noisy queries, and $q_{N_{ij}}\in\mathbb{R}^{K \times D}$, with $K$ signifying the number of objects present in the input image. These noisy queries are generated by a Variational Query Generator to overcome the gradient vanishing issue associated with conventional denoising. During training, the predictions derived from the learnable queries are passed to 3D-Aware Bipartite Matching, which addresses the mismatch issues often encountered with naive 2D-only approaches.
\subsection{3D-Aware Bipartite Matching}
\label{sec:method:3D_bipartite_matching}
\textbf{3D Matching.} To correctly associate each query with its corresponding ground-truth object, we introduce a novel matching cost function, which is formally defined as:
\begin{equation}
    C_{match}=C_{2D} + \Gamma(t) C_{3D}
    \label{eq:matching_eq}
\end{equation}
The total matching cost is a weighted combination of 2D and 3D prediction costs. Specifically, $C_{2D}$ aggregates costs related to the object category, 2D bounding box size, and the projected 3D center. Conversely, $C_{3D}$ encompasses costs for depth, 3D size, and orientation angle. $\Gamma(t) \in [0, 1]$ serves as a scheduler weight. This weight is designed to be low during the initial training phase when 3D predictions are inherently unstable, and it gradually increases as the 3D estimation becomes more reliable.

\noindent\textbf{3D Weight Scheduler.} During initial training, the instability of the predicted 3D attributes makes their immediate integration into the Bipartite Matching problematic. We address this unreliability by introducing a step scheduler. This scheduler controls and delays the integration of the 3D attributes until prediction stability improves, which is formally defined as:
\begin{equation}
      \Gamma(t)=
      \left\{
      \begin{array}{l}
           0, \texttt{ if } t < T \\
           \epsilon, \texttt{ otherwise}
      \end{array}
      \right.
      \label{eq:3d_weight_scheduler}
\end{equation}
where $t$ denotes the current training epoch, $\epsilon$ is the weight threshold and $T$ is the trigger epoch, described detailed in the supplementary. 
\subsection{3D-DeNoising}
\label{sec:method:3d_denoising}
The use of a scheduler weight, $\Gamma(t)$, when integrating 3D attributes into the Bipartite Matching process is designed to mitigate the mismatch problem in 2D-only approach. However, the potential benefit is severely compromised by the instability introduced by the ill-posed nature of 3D estimation from monocular images. This fundamental limitation restricts the method's impact, leading to performance that is only marginally better than the 2D-only approach, as shown in \ref{tab:contribution_ablation}(c). To overcome this limitation and decisively stabilize the 3D integration, we propose 3D-Denoising. By attaching the 3D ground truth to the noisy query generator, we inject strong, reliable 3D supervision directly into the training phase. This supervision acts as a robust anchor, guiding the network to learn a more stable and accurate 3D representation despite the noisy inputs, thereby stabilizing the benefits of the 3D attribute integration into the Bipartite Matching.

\noindent\textbf{3D Noisy Query.} In 2D object detection, DN-DETR~\cite{DN-DETR} utilizes the bounding box as the reference anchor and category of object to generate the noise query. We first reformulate how to generate a noisy query to enhance the effect of integrating 3D-attribute into Bipartite Matching. The objective of monocular 3D object detection consist of: category $c$, projected center $\left(x_c,y_c\right)$, 2D bounding box $l,r,t,b$, 3D dimension $l_{3D},w_{3D},h_{3D}$, orientation $\theta$ and central depth $d$. After generating noise boxes, we denote a 6D anchor box $\left(x_c,y_c,l,r,t,b\right)$ as the initial reference. Then we map the 3D information $\left(c,l_{3D},w_{3D},h_{3D},\theta,d\right)$ to a hidden space using an embedding layer, yielding the 3D query. The anchor box and the query are fed into the decoder to be reconstructed. The specific details of the embedding layer and the noise generation mechanism are provided in the supplementary material.

\noindent\textbf{Mask Separated Self-Attention.} To improve detection accuracy, the baseline MonoDETR model incorporates a group-wise one-to-many assignment strategy~\cite{GroupDETR}, employing $G$ distinct groups of learnable queries $Q_L = \{q_{L_i}\}_{i=1}^G$. This approach utilizes Separate Self-Attention~\cite{GroupDETR} to ensure that queries belonging to different groups do not interact, thereby maintaining group independence. To leverage the benefits of both the denoising training paradigm and the group-wise assignment methodology, we introduce the Mask Separated Self-Attention mechanism. This mechanism employs a predefined attention $M \in \{0,1\}^{S \times S}$ inspired by DN-DETR~\cite{DN-DETR}, where $S = K \cdot C + N$, to regulate the interactions only among noisy queries, as well as between noisy queries and learnable queries. Specifically, from the sets of learnable queries $Q_L$ and noisy queries $Q_N$, we construct a new set of combined queries $Q = \{q_i\}_{i=1}^G$, where each $q_i \in \mathbb{R}^{S \times D}$ is constructed by concatenating the learnable query $q_{L_i}$ with its corresponding noisy queries $\{q_{N_{ij}}\}_{j=1}^C$:
\begin{equation}
    q_i = \text{concat}( q_{N_{i1}}, q_{N_{i2}}, ..., q_{N_{iC}},q_{L_i})
    \label{eq:q_i}
\end{equation}
This newly formed set of queries is then processed by Separated Self-Attention with the predefined mask $M$. Through this proposed Mask Separated Self-Attention, we achieve precise control over the interaction patterns between queries of different types and within each group of queries.
\begin{figure*}[h!]
    \centering
    % \includesvg[width=\textwidth]{fig/motivation_vdn.svg}
    \includegraphics[width=\textwidth]{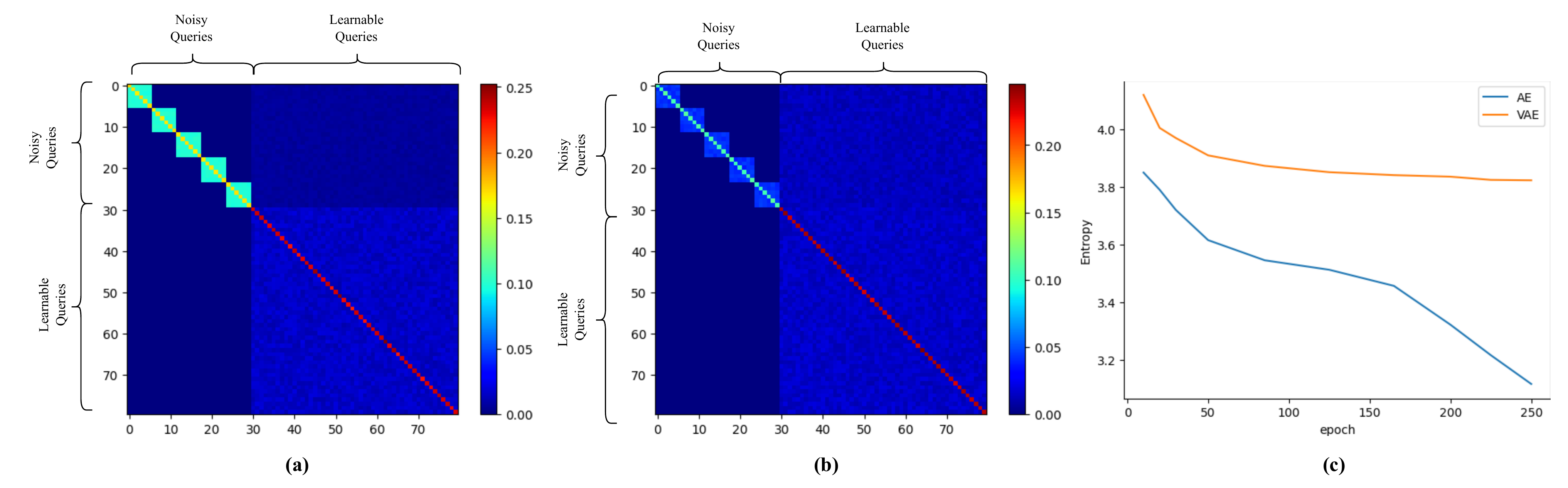}
    \caption{\textbf{The influence of the denoised query on learnable queries.} The self-attention maps trend resulting from the conventional denoising method and the proposed Variational Query DeNoising approach are illustrated in (a) and (b), respectively. (c) presents the average entropy of the attention maps for both methods throughout the entire training period.}
    \label{fig:motivation_vdn}
\end{figure*}
\subsection{Variational Query DeNoising}
\label{sec:method:variational_query_denoising}
\textbf{Challenges with Conventional DeNoising.} While the integration of conventional denoising techniques can stabilize the integration of 3D-attribute into the Bipartite Matching process and offers an initial improvement in model detection performance as shown in ~\cref{tab:contribution_ablation}(e), this approach encounters the gradient vanishing problem. We can visually demonstrate this issue by inspecting the attention map \(A_i \in \mathbb{R}^{S \times S}\) derived from the Mask Separated Self-Attention mechanism for the \(i\)-th group of learnable queries (Fig.~\ref{fig:motivation_vdn}a). Observation of the attention map reveals a significant decoupling effect. Specifically, the attention scores between the noisy queries and the learnable queries approach zero, visibly in the upper-right quadrant of the map. This decoupling is detrimental because it obstructs the efficient backpropagation of gradients originating from the reconstruction loss to the learnable queries, thereby limiting their capacity for further adaptation and improvement.

To quantitatively analyze this behavior, we monitored the sparsity of the attention maps \(A_i\) during the evaluation of Mono3DV across training epochs. Treating attention maps as probability distributions, we employed negative entropy as an intuitive metric for sparsity (a lower entropy indicates greater sparsity). Increased sparsity signifies that noisy queries are predominantly attending to themselves rather than the learnable queries. This elevated self-attention among noisy queries directly diminishes the attention weight (and consequently, the gradient flow) directed towards the learnable queries. As illustrated in Fig.~\ref{fig:motivation_vdn}c, the attention map associated with the conventional denoising method (AE) exhibits a rapid decrease in negative entropy (increase in sparsity) throughout training, providing further evidence of the restricted gradient flow and resulting performance plateau.

\noindent\textbf{Variational Query DeNoising.} To overcome the identified gradient vanishing problem and enhance the effectiveness of the denoising process, we propose leveraging the inherent stochastic properties of a Variational Autoencoder (VAE). Our approach aims to introduce beneficial variation into the noisy queries, which in turn increases the entropy of the attention distribution and promotes more robust gradient propagation.

As depicted in the overall architecture shown in Fig.~\ref{fig:overall} (Variational Query Generator), the process begins by feeding the initial noisy input boxes through a dedicated box embedding layer. This layer functions as the encoder of our VAE, predicting $\mu$ and $\Sigma$ parameters of a latent distribution. The stochastic noisy queries are then synthesized by sampling from this learned distribution using the reparameterization trick: \(z \sim \mathcal{N}\left(\mu,\Sigma\right)\). These generated stochastic queries are subsequently processed by the model's decoder, analogous to how queries are handled in a standard denoising setup. Training for this denoising process is guided by a denoising loss function, defined as:
\begin{equation}
    L_{DN} = L_{res} + \beta L_{KL}\left(\mathcal{N}\left(\mu,\Sigma\right),\mathcal{N}\left(0,I\right)\right)
    \label{eq:vdn_loss}
\end{equation}
with $L_{res}$ representing reconstruction loss computed from the noisy queries, $L_{KL}$ denotes the Kullback-Leibler divergence loss that regularizes the learned distribution towards a standard normal prior $\mathcal{N}(0,I)$, and $\beta$ is a weighting factor.

~\cref{fig:motivation_vdn}c demonstrates that the proposed Variational Query Denoising (VAE) maintains significantly higher attention map entropy compared to the conventional denoising method, indicating reduced sparsity. Furthermore, as shown in ~\cref{fig:motivation_vdn}b (contrast with ~\cref{fig:motivation_vdn}a), the attention scores in the upper-right corner of the self-attention map do not converge to zero. This lack of decoupling confirms that the learnable queries continue to effectively interact with and benefit from the denoising process enabled by our variational approach.
\subsection{Loss Function}
\label{sec:method:loss_function}
\textbf{Forward-Looking Distillation.} We propose that integrating self-distillation enhances the iterative refinement strategy in DETR~\cite{DETR} by transferring knowledge from the final high-performing decoder layer to shallower layers. To prioritize high-quality predictions, we weight the distillation loss using the $\mathbf{IoU}_{3D}$ between the last decoder prediction and the ground-truth 3D box. Following ~\cite{SD} success, a shared MLP $\left(f_Q\left(\cdot\right)\right)$ refines the query to improve the quality of the distilled knowledge. The self-distillation loss is formally expressed as:
\begin{equation}
    L_{dis} = \sum_{i=1}^{D-1}{\mathbf{IoU}_{3D}}_{Q_D}\cdot \texttt{Smooth\_L1}\left(f_Q\left(Q_i\right), Q_D\right)
    \label{eq:loss_dis}
\end{equation}
where $Q_i$ represents the output query of the i-th decoder layer, $D$ denotes the total number of decoder layers in the model. It is important to note that Forward-Looking Distillation is applied to both the learnable queries $Q_L$ and the noisy queries $Q_N$.

\begin{table*}[ht!]
\centering
\caption{Comparisons with state-of-the-art monocular methods on the KITTI test and val sets for the car category. We \textbf{bold} the best results and \underline{underline} the second-best results. The {\color[HTML]{3166FF} blue} refers to the gain and the {\color[HTML]{FE0000} red} is the decrease of our method.}
\resizebox{\textwidth}{!}{%\textbf{
\begin{tabular}{l|c|c|clcc|ccc|clcc|clcc}
\hline
\multicolumn{1}{c|}{} & & & \multicolumn{4}{c|}{Test, $AP_{3D|R40}$} & \multicolumn{3}{c|}{Test, $AP_{BEV|R40}$} & \multicolumn{4}{c|}{Val, $AP_{3D|R40}$} & \multicolumn{4}{c}{Val, $AP_{BEV|R40}$} \\ \cline{4-18}
\multicolumn{1}{l|}{\multirow{-2}{*}{Methods}} & \multirow{-2}{*}{Extra data} & \multirow{-2}{*}{Reference} & \multicolumn{2}{c}{Easy} & Mod. & Hard & Easy & Mod. & Hard & \multicolumn{2}{c}{Easy} & Mod. & Hard & \multicolumn{2}{c}{Easy} & Mod. & Hard \\ \hline
MonoDTR~\cite{MonoDTR} & & CVPR 2022 & \multicolumn{2}{c}{21.99} & 15.39 & 12.73 & 28.59 & 20.38 & 17.14 & \multicolumn{2}{c}{24.52} & 18.57 & 15.51 & \multicolumn{2}{c}{33.33} & 25.35 & 21.68 \\
DID-M3D~\cite{DID-M3D} &  & ECCV 2022 & \multicolumn{2}{c}{24.40} & 16.29 & 13.75 & 32.95 & 22.76 & 19.83 & \multicolumn{2}{c}{22.98} & 16.12 & 14.03 & \multicolumn{2}{c}{31.10} & 22.76 & 19.50 \\
OccupancyM3D~\cite{OccupanyM3D} & \multirow{-3}{*}{LiDAR} & CVPR 2024 & \multicolumn{2}{c}{25.55} & 17.02 & 14.79 & \underline{35.38} & 24.18 & \underline{21.37} & \multicolumn{2}{c}{26.87} & 19.96 & 17.15 & \multicolumn{2}{c}{35.72} & 26.60 & 23.68 \\ \hline
MonoPGC~\cite{MonoPGC} &  & ICRA 2023 & \multicolumn{2}{c}{24.68} & 17.17 & 14.14 & 32.50 & 23.14 & 20.30 & \multicolumn{2}{c}{25.67} & 18.63 & 15.65 & \multicolumn{2}{c}{34.06} & 24.26 & 20.78 \\
OPA-3D~\cite{OPA-3D} & \multirow{-2}{*}{Depth} & RAL 2023 & \multicolumn{2}{c}{24.60} & 17.05 & 14.25 & 33.54 & 22.53 & 19.22 & \multicolumn{2}{c}{24.97} & 19.40 & 16.59 & \multicolumn{2}{c}{33.80} & 25.51 & 22.13 \\ \hline
MonoCon~\cite{MonoCon} &  & AAAI 2022 & \multicolumn{2}{c}{22.50} & 16.46 & 13.95 & 31.12 & 22.10 & 19.00 & \multicolumn{2}{c}{26.33} & 19.01 & 15.98 & \multicolumn{2}{c}{-} & - & - \\
DEVIANT~\cite{DEVIANT} &  & ECCV 2022 & \multicolumn{2}{c}{21.88} & 14.46 & 11.89 & 29.65 & 20.44 & 17.43 & \multicolumn{2}{c}{24.63} & 16.54 & 14.52 & \multicolumn{2}{c}{32.60} & 23.04 & 19.99 \\
MonoDDE~\cite{MonoDDE} &  & CVPR 2022 & \multicolumn{2}{c}{24.93} & 17.14 & 15.10 & 33.58 & 23.46 & 20.37 & \multicolumn{2}{c}{26.66} & 19.75 & 16.72 & \multicolumn{2}{c}{35.51} & 26.48 & 23.07 \\
MonoUNI~\cite{MonoUNI} &  & NeurlPS 2023 & \multicolumn{2}{c}{24.75} & 16.73 & 13.49 & - & - & - & \multicolumn{2}{c}{24.51} & 17.18 & 14.01 & \multicolumn{2}{c}{-} & - & - \\
MonoDETR~\cite{MonoDETR} &  & ICCV 2023 & \multicolumn{2}{c}{25.00} & 16.47 & 13.58 & 33.60 & 22.11 & 18.60 & \multicolumn{2}{c}{28.84} & 20.61 & 16.38 & \multicolumn{2}{c}{37.86} & 26.95 & 22.80 \\
MonoCD~\cite{MonoCD} &  & CVPR 2024 & \multicolumn{2}{c}{25.53} & 16.59 & 14.53 & 33.41 & 22.81 & 19.57 & \multicolumn{2}{c}{26.45} & 19.37 & 16.38 & \multicolumn{2}{c}{34.60} & 24.96 & 21.51 \\
FD3D~\cite{FD3D} & & AAAI 2024 & \multicolumn{2}{c}{25.38} & 17.12 & 14.50 & 34.20 & 23.72 & 20.76 & \multicolumn{2}{c}{28.22} & 20.23 & 17.04 & \multicolumn{2}{c}{36.98} & 26.77 & 23.16 \\
MonoDGP~\cite{MonoDGP} & \multirow{-8}{*}{None} & CVPR 2025 & \multicolumn{2}{c}{\underline{26.35}} & \underline{18.72} & \underline{15.97} & 35.24 & \textbf{25.23} & \textbf{22.02} & \multicolumn{2}{c}{\underline{30.76}} & \underline{22.34} & \underline{19.02} & \multicolumn{2}{c}{\underline{39.40}} & \underline{28.20} & \underline{24.42} \\ \hline
Mono3DV (Ours) & None & - & \multicolumn{2}{c}{\textbf{28.26}} & \textbf{19.20} & \textbf{16.21} & \textbf{35.77} & \underline{24.82} & \underline{21.37} & \multicolumn{2}{c}{\textbf{32.12}} & \textbf{23.55} & \textbf{20.15} & \multicolumn{2}{c}{\textbf{40.85}} & \textbf{29.24} & \textbf{25.49} \\
Improvement & - & - & \multicolumn{2}{c}{{\color[HTML]{3166FF} +1.91}} & {\color[HTML]{3166FF} +0.48} & {\color[HTML]{3166FF} +0.24} & {\color[HTML]{3166FF} +0.39} & {\color[HTML]{FE0000} -0.41} & {\color[HTML]{FE0000} -0.65} & \multicolumn{2}{c}{{\color[HTML]{3166FF} +1.36}} & {\color[HTML]{3166FF} +1.21} & {\color[HTML]{3166FF} +1.13} & \multicolumn{2}{c}{{\color[HTML]{3166FF} +1.45}} & {\color[HTML]{3166FF} +1.04} & {\color[HTML]{3166FF} +1.07} \\ \hline
\end{tabular}}
\label{tab:main_result}
\end{table*}

\noindent\textbf{Overall Loss.} The training loss of Mono3DV is composed of three distinct terms: detection loss $L_{det}$, denoising loss $L_{DN}$, and self-distillation loss $L_{dis}$. We adopt $L_{det}$ formulation from MonoDETR~\cite{MonoDETR} that includes losses for object category, projected center point, 2D bounding box, orientation, 3D size, central depth, and depth map. In addition, we propose the denoising loss, $L_{DN}$ derived from Variational Query DeNoising as presented in \cref{eq:vdn_loss}, and $L_{dis}$ is integrated for the proposed Forward-Looking Distillation as mentioned in \cref{eq:loss_dis}. The overall loss of Mono3DV is then formulated as:
\begin{equation}
    L_{overall} = \lambda_1L_{det}+ \lambda_{2}L_{DN}+ \lambda_{3}L_{dis}
    \label{eq:overall_loss}
\end{equation}
\section{Experiments}
\label{sec:experiments}
\subsection{Settings}
\textbf{Dataset.} Our model was evaluated on the widely-used KITTI 3D object detection benchmark~\cite{KITTIDATA}. This dataset contains 7481 training images and 7581 testing images. Following the methodology of Chen et al.~\cite{Chen20153DOP}, we split the training set into two subsets: a training set of 3712 images and a validation set of 3769 images. This split facilitated ablation studies to assess the effectiveness of different components within our Mono3DV model.

\noindent\textbf{Evaluation metrics.} We evaluated the detection performance in three difficulty levels: easy, moderate, and hard. We used two primary metrics: $AP_{3D}$ and $AP_{BEV}$ indicate the accuracy of 3D bounding box predictions and the accuracy of 2D projections of 3D bounding boxes onto a bird's-eye view respectively. Both $AP_{3D}$ and $AP_{BEV}$ were calculated at 40 recall positions~\cite{simonelli2019disentangling}.

\noindent\textbf{Implementation Details.} Our network utilizes the ResNet50~\cite{ResNet} as its backbone and was trained for 250 epochs using the Adam optimizer with a batch size of 8 images on a single NVIDIA 3090 GPU. The learning rate was initialized at 0.0002 and decayed by a factor of 0.5 at epochs 85, 125, 165, and 225. The weights of losses are set as $\{1, 1, 0.5\}$ for $\lambda_1$ to $\lambda_3$. During the inference, queries with a category confidence below 0.2 are discarded.
\label{sec:experiments:setttings}
\subsection{Main Results}
\label{sec:experiments:main_results}
\textbf{Experiment on the KITTI 3D test set.}
As shown in \cref{tab:main_result}, our Mono3DV method demonstrates superior performance on the KITTI test set compared to state-of-the-art monocular 3D object detection methods. Specifically, Mono3DV achieves a significant improvement in $\mathbf{AP_{3D}}$, surpassing the second-best method by $+1.91\%$, $+0.48\%$, and $+0.24\%$ across the three difficulty levels. Furthermore, it outperforms the second-best in $\mathbf{AP_{BEV}}$ by $0.39\%$ under the easy difficulty while maintaining competitive performance for moderate and hard cases. These results underscore the effectiveness of our proposed framework for accurate 3D object prediction from monocular images.

\noindent\textbf{Experiment on the KITTI 3D val set.} We also evaluated our approach on the KITTI validation dataset. As presented in \cref{tab:main_result}, our Mono3DV demonstrates superior performance compared to all existing methods. Notably, it surpasses the second-best approach under three-level difficulties by $+1.36\%$, $+1.21\%$, and $+1.13\%$ in $\mathbf{AP_{3D}}$, and by $+1.45\%$, $+1.04\%$, and $+1.07\%$ in $\mathbf{AP_{BEV}}$. These results further emphasize the effectiveness of Mono3DV.

\noindent\textbf{Efficiency.}
As shown in Table \cref{tab:efficiency}, our proposed Mono3DV maintains the same computational budget as the efficient baseline MonoDETR~\cite{MonoDETR} while achieving a substantial performance boost in $AP_{3D}$. Furthermore, compared to the second-best method MonoDGP~\cite{MonoDGP}, Mono3DV demonstrates clear dominance by outperforming it in both 3D detection accuracy and inference efficiency.
\begin{table}[h!]
\centering
\caption{\textbf{Efficiency comparison.} We test the Runtime (ms) on one RTX 3090 GPU with batch size 1, and compare $AP_{3D}$ on \textit{test} set.}
\resizebox{\columnwidth}{!}{
	\begin{tabular}{l c c c} % Transposed columns: 4 total
	\toprule
	Method & Runtime$\downarrow$ & GFlops$\downarrow$ & $AP_{3D}$ Mod.\\ % Metric headers
	\midrule
	MonoDTR~\cite{MonoDTR} & \textbf{37} & 120.48 & 15.39\\
	MonoDETR~\cite{MonoDETR} & 38 & \textbf{62.12} & 16.47\\
	MonoDGP~\cite{MonoDGP} & 45 & 71.76 & 18.72\\
	Mono3DV & 38 & \textbf{62.12} & \textbf{19.20}\\
	\bottomrule
	\end{tabular}}
\vspace{0.05cm}
\label{tab:efficiency}
\end{table}
\subsection{Ablation Study}
\label{sec:experiments:ablation_study}
We verify the effectiveness of each of our components and report in $\mathbf{AP_{3D}}$ for the \textit{Car} category on the KITTI val set.

\begin{table}[h!]
    \centering
    \caption{\textbf{Analysis of different components of our approach} on the Car category of the KITTI validation set. `FLD' denotes the Forward-Looking Distillation. `3DM' denotes the 3D-Aware Bipartie Matching.`3DN' denotes 3D-DeNoising, and `VDN' denotes the Variational Query DeNoising.}
    \resizebox{\columnwidth}{!}{
    \begin{tabular}{c|cccc|ccc}
         \toprule
         \multicolumn{1}{l|}{\multirow{2}{*}{}} &
         \multicolumn{4}{c|}{Ablation} & \multicolumn{3}{c}{Val, $AP_{3D|R40}$} \\
         & FLD & 3DM & 3DN & VDN & Easy & Mod. & Hard \\ \midrule \midrule
         (a) & \usym{2715} & \usym{2715} & \usym{2715} & \usym{2715} & 29.99 & 20.92 & 17.44 \\
         (b) & \Checkmark & \usym{2715} & \usym{2715} & \usym{2715} & 30.05 & 21.54 & 18.31 \\
         (c) & \Checkmark & \Checkmark & \usym{2715} & \usym{2715} & 30.09 & 21.63 & 18.27 \\
         (d) & \Checkmark & \usym{2715} & \Checkmark & \usym{2715} & 30.06 & 21.59 & 18.34 \\
         (e) & \Checkmark & \Checkmark & \Checkmark & \usym{2715} & 30.46 & 22.78 & 19.49\\
         (f) & \Checkmark & \Checkmark & \Checkmark & \Checkmark & \textbf{32.12} & \textbf{23.55} & \textbf{20.15}\\ \midrule
         \multicolumn{5}{c|}{Improvement} & +2.13 & +2.63 & +2.71\\ 
         \bottomrule
    \end{tabular}}
    \label{tab:contribution_ablation}
\end{table}
\noindent\textbf{Effectiveness of each proposed component.} In \cref{tab:contribution_ablation}, we present an ablation study analyzing the effectiveness of our proposed components. The evaluation begins with (a) The baseline, MonoDETR~\cite{MonoDETR}, which lacks the proposed enhancement modules. We then individually assess the contribution of (b) Forward-Looking Distillation, which is integrated to enhance the iterative refinement strategy, and (c) 3D-Aware Bipartite Matching, which is designed to resolve the mismatch problem of the 2D-only approach. Furthermore, we investigate the effect of (d) Adding 3D-Denoising to the training phase through Mask Separated Self-Attention. A combined model evaluates the benefit of integrating (e) both 3D-Aware Bipartite Matching and 3D-Denoising to mitigate instability caused by ill-posed 3D estimation from monocular images. Finally, we assess the impact of (f) Variational Query Denoising, which enhances the denoising effect on the set of learnable queries.

Firstly, (b) Forward-Looking Distillation significantly improves detection quality, particularly in challenging scenarios (moderate and hard settings), achieving gains of $\mathbf{+0.62\%,+0.87\%}$ in $\mathbf{AP_{3D}}$, respectively. This validates the efficacy of our proposed self-distillation approach. Secondly, incorporating (c) 3D-Aware Bipartite Matching alone does not significantly enhance model performance. Thirdly, adding (d) 3D-Denoising results in performance nearly identical to (b). Fourthly, although separate integration of 3D-Aware Matching (c) and 3D-Denoising (d) does not lead to substantial gains, combining them (e) yields a significant performance increase. Compared to (b), the combined model achieves gains of $\mathbf{0.41\%, 1.24\%, 1.18\%}$ across the three difficulty levels in $\mathbf{AP_{3D}}$. Finally, integrating (f) Variational Query Denoising yields substantial improvements across all difficulty levels compared to (e), achieving gains of $\mathbf{+1.66\%, +0.77\%, +0.66\%}$, respectively. This significant improvement underscores the effectiveness of incorporating stochastic properties into the denoising process.

\begin{table}[h!]
    \centering
    \caption{\textbf{Ablation study on the design of 3D Weight Scheduler $\Gamma(t)$} in 3D-Aware Bipartite Matching.}
    \begin{tabular}{c|ccc}
         \toprule
         \multirow{2}{*}{Ablation} & \multicolumn{3}{c}{Val, $AP_{3D|R40}$} \\
          & Easy & Mod. & Hard \\ \midrule \midrule
          $\Gamma(t)=0$ & 30.05 & 21.54 & \textbf{18.31} \\
          $\Gamma(t)=1$ & 3.26 & 1.12 & 0.07 \\
          $\Gamma(t)=0.1$ & 21.15 & 17.21 & 13.42 \\
          HTL~\cite{GUPNet} & 29.22 & 21.53 & 18.15 \\
          Step & \textbf{30.09} & \textbf{21.63} & 18.27 \\
         \bottomrule
    \end{tabular}
    \label{tab:3d_weight_scheduler}
\end{table}
\noindent\textbf{3D-Aware Bipartite Matching.} We explore multiple strategies for the 3D weight scheduler $\Gamma(t)$ when integrating 3D attributes into the Bipartite Matching process. As detailed in \cref{tab:3d_weight_scheduler}, our initial design of $\Gamma(t)$ as a constant showed a rapid degradation in performance as the constant increased. Notably, setting $\Gamma(t)=1$ severely disrupted the training process, preventing model convergence. While a smaller value of $\Gamma(t)=0.1$ allowed the model to converge, its performance was still poor compared to the baseline 2D-only Bipartite Matching ($\Gamma(t)=0$). We also implemented a Hierarchical Task Learning (HTL) \cite{GUPNet}, which yielded more stable results and achieved approximately comparable performance. Finally, our proposed step scheduler demonstrated an improvement in performance over the baseline, further emphasizing the efficacy of the proposed 3D-Aware Bipartite Matching.

\begin{figure*}[h!]
    \centering
    \includegraphics[width=\textwidth]{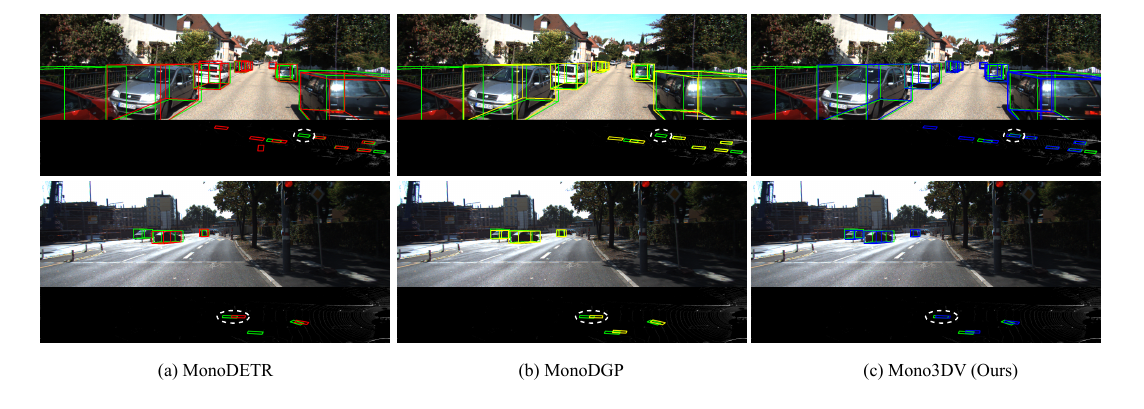}
    \caption{\textbf{Qualitative results on KITTI val set.} (a) MonoDETR~\cite{MonoDETR}. (b) MonoDGP~\cite{MonoDGP}. (c) Mono3DV (Ours). For each image set, the top row presents the camera-view visualization, while the bottom row offers the corresponding bird’s-eye view. Ground-truth bounding boxes are rendered in \textcolor{green}{green}, and predictions are shown in order: \textcolor{red}{red}, \textcolor{yellow}{yellow}, \textcolor{blue}{blue}. We also circle some objects to highlight the difference between other state of the art and our method.}
\label{fig:qualitative_results}
\end{figure*}
\begin{table}[ht!]
    \centering
    \caption{\textbf{Ablation study on the effect of ground truth} in generating a noised query. `Category' denotes only using category to generate the noised query. `3DBox' denotes using the 3D boxes.}
    \begin{tabular}{cc|ccc}
         \toprule
         \multicolumn{2}{c|}{Ablation} & \multicolumn{3}{c}{Val, $AP_{3D|R40}$} \\
         Category& 3DBox & Easy & Mod. & Hard \\ \midrule \midrule
         \usym{2715} & \usym{2715} &  30.09 & 21.63 & 18.27 \\
         \checkmark & \usym{2715}  & 29.16 & 21.28 & 18.00\\
         \checkmark & \checkmark & \textbf{30.46} & \textbf{22.78} & \textbf{19.49}\\
         \bottomrule
    \end{tabular}
    \label{tab:noise_query_generator_ablation}
\end{table}
\noindent\textbf{3D Noisy Query.} To investigate the impact of query representation on denoising performance, we conducted ablation studies using different query types, summarized in \cref{tab:noise_query_generator_ablation}. Initially, we followed DN-DETR~\cite{DN-DETR} and employed only category embeddings as queries. However, this approach resulted in a slight performance degradation. We hypothesized that using category information alone might be insufficient to accurately represent 3D objects, consequently reducing its effectiveness in stabilizing the Bipartite Matching process. To address this limitation, we incorporated 3D bounding box information into the query representation, as detailed in \cref{sec:method:3d_denoising}. This enhancement led to a significant improvement in the overall performance.

\begin{table}[h!]
    \centering
    \caption{\textbf{Ablation study on the effect of $\beta$} in Variational Query DeNoising.}
    \begin{tabular}{c|ccc}
         \toprule
         \multirow{2}{*}{Ablation} & \multicolumn{3}{c}{Val, $AP_{3D|R40}$} \\
          & Easy & Mod. & Hard \\ \midrule \midrule
          $\beta=0$ & 30.46 & 22.78 & 19.49 \\
          $\beta = 1.0$ & 30.42 & 22.81 & 19.32 \\
          $\beta = 4.0$ & \textbf{32.12} & \textbf{23.55} & \textbf{20.15} \\
          $\beta = 10.0$ & 29.98 & 21.32 & 18.62 \\
         \bottomrule
    \end{tabular}
    \label{tab:beta_vdn_ablation}
\end{table}
\noindent\textbf{Variational Query DeNoising} For the Variational Query DeNoising, we compare the model's performance with different KL-divergence weights $\beta$. Initially, we set the $\beta$ hyperparameter to 1.0. As shown in~\cref{tab:beta_vdn_ablation}, the performance under this configuration was only marginally different from the original denoising approach. This suggests the influence of the KL-divergence loss term was insufficient, causing the model to behave similarly to a conventional denoising autoencoder. Following the principles of $\beta$-VAE~\cite{betaVAE}, we subsequently increased $\beta$ to 4.0. This modification significantly enhanced the model's performance. Moreover, we expanded the experiment with $\beta=10.0$. However, its performance was degenerate, likely due to an overly dominant KL-divergence term leading to posterior collapse.

\subsection{Qualitative Results}
Qualitative visualizations, shown in \cref{fig:qualitative_results}, allow for an intuitive comparison of our approach against the baseline. Compared to other state-of-the-art methods, Mono3DV demonstrates improved localization accuracy across most objects in the scene. The integration of 3D matching is key to this performance boost, enabling the monocular 3D detector to achieve more accurate localization.
\section{Conclusion}
\label{sec:conclusion}
In this work, we proposed Mono3DV, a novel transformer-based method for monocular 3D object detection that overcomes the mismatch problem in 2D-only Bipartite Matching and the training instability caused by naive 3D attribute integration. Our core innovation is the 3D-Aware Bipartite Matching, which effectively incorporates the crucial 3D attribute into the matching process using a stabilizing scheduler, thereby resolving the mismatch problems inherent in naive 2D-only strategies. To further enhance robustness against the ill-posed nature of monocular 3D estimation, we introduced 3D DeNoising to stabilize the Bipartite Matching. Finally, recognizing the challenge of gradient vanishing in conventional denoising, we presented Variational Query DeNoising, which significantly boosted the model's performance. Evaluated on the demanding KITTI 3D object detection benchmark, Mono3DV achieves state-of-the-art performance among all monocular detectors without leveraging any additional training data, confirming the efficacy of our architectural and methodological contributions.
\clearpage
{
    \small
    \bibliographystyle{ieeenat_fullname}
    \bibliography{main}
}
\clearpage
\setcounter{page}{1}
\appendix
\maketitlesupplementary

\section{Additional Implementation Details}
\subsection{3D-Aware Bipartite Matching}
Recall that the matching cost of the 3D-Aware Bipartite Matching is defined as:
\begin{equation}
    C_{match}=C_{2D} + \Gamma(t) C_{3D}
    \label{eq:matching_eq_recall}
\end{equation}
with $C_{2D}$ is the 2D cost, $C_{3D}$ denotes the 3D cost, and $\Gamma(t)$ denotes the scheduler weight.

We follow the baseline MonoDETR~\cite{MonoDETR} and constructed the 2D cost as:
\begin{equation}
    \begin{aligned}
        C_{2D} & =\lambda_{cls}C_{cls} + \lambda_{proj}C_{xy3D} \\
        & \hspace{2em} + \lambda_{lrtb}C_{lrtb} + \lambda_{GIoU}C_{GIoU}
    \end{aligned}
\end{equation}
with $\lambda_{cls}=2, \lambda_{proj}=10, \lambda_{lrtb} = 5, \lambda_{GIoU} = 2$.

The 3D cost was formulated by defining a cost for each of the three 3D attributes, specifically size, angle, and depth, the detailed formulation of which is defined as follows:
\begin{equation}
    C_{3D} = C_{size3D} + C_{orien} + C_{depth}
\end{equation}
\subsection{3D Noisy Query}
To begin the denoising process, we first generate noisy bounding box information derived from the original ground truth, encompassing the projected 3D center $(x_c, y_c)$, 2D bounding box coordinates $\left(l,r,t,b\right)$, 3D object dimensions ($l_{3D}$, $h_{3D}$, $w_{3D}$), object depth $d$, category label $c$, and orientation $\theta$. Subsequently, these noisy boxes serve as input to the embedding layer, producing noisy queries. Furthermore, we introduce parameters $\lambda_C$ and $\lambda_D$ to manage the extent of applied noise.

\noindent\textbf{Projected 3D Center} $\left(x_c,y_c\right)$. Inspired by DN-DETR~\cite{DN-DETR}, we perform \textbf{center shifting} on the projected 3D center so that it still lies inside the original 2D bounding box by following the formula:
\begin{equation}
     \begin{aligned}
         \widetilde{x}_c &= x_c + \lambda_C \mathcal{U}(-1, 1) \cdot \frac{r+l}{2} \\
         \widetilde{y}_c &= y_c + \lambda_C \mathcal{U}(-1, 1) \cdot \frac{t+b}{2}
     \end{aligned}
\end{equation}
\noindent\textbf{2D Size} $\left(l,r,t,b\right)$. We also perform \textbf{box scaling} on the 2D bounding box, by manually adjusting $l,r,t,b$ through following formulation:
\begin{equation}
    \begin{aligned}
        \widetilde{l} &= \texttt{clip}\left(l + \lambda_C \mathcal{U}(-1, 1) \cdot l, 0, 1\right) \\
        \widetilde{r} &= \texttt{clip}\left(l + \lambda_C \mathcal{U}(-1, 1) \cdot r, 0, 1\right) \\
        \widetilde{t} &= \texttt{clip}\left(l + \lambda_C \mathcal{U}(-1, 1) \cdot t, 0, 1\right) \\
        \widetilde{b} &= \texttt{clip}\left(l + \lambda_C \mathcal{U}(-1, 1) \cdot b, 0, 1\right) \\
    \end{aligned}
\end{equation}
with \texttt{clip()} operation to limit the noisy value in range $\left[0,1\right]$.

\noindent\textbf{3D Size} $\left(l_{3D},h_{3D},w_{3D}\right)$. Inspired by established 2D bounding box scaling techniques, we extend this concept to 3D dimensions, employing the following formulation:
\begin{equation}
    \begin{aligned}
        \widetilde{l}_{3D} &= l_{3D} + \lambda_C \mathcal{U}(-1, 1) \cdot l_{3D} \\
        \widetilde{h}_{3D} &= h_{3D} + \lambda_C \mathcal{U}(-1, 1) \cdot h_{3D} \\
        \widetilde{w}_{3D} &= w_{3D} + \lambda_C \mathcal{U}(-1, 1) \cdot w_{3D}
    \end{aligned}
\end{equation}
\noindent\textbf{Object Depth} $d$. Noisy depth is generated by randomly adjusting the center depth along the object's length, with the condition that the resulting noisy depth value stays within the 3D bounding box, formulated as:

\begin{equation}
    \begin{aligned}
        \widetilde{d} &= d + \lambda_C \mathcal{U}
        (-1, 1) \cdot \frac{l_{3D}}{2}
    \end{aligned}
\end{equation}

\noindent\textbf{Category} $c$. By adopting label flipping~\cite{DN-DETR} for label noising with flipping rate $\lambda_D$, we aim to improve the model's ability to learn the relationship between object labels and their corresponding noisy bounding box predictions.

\noindent\textbf{Orientation} $\theta$. The baseline MonoDETR~\cite{MonoDETR} represents orientation by dividing it into multiple discrete bins, denoted as $\theta_{bin}$, and a continuous residual, denoted as $\theta_r$. To introduce noise into the orientation, we retain the residual $\theta_r$ and perform bin flipping with flipping rate $\lambda_D$. This involves randomly assigning the ground truth orientation bin to a different bin. Subsequently, only this noisy orientation bin is fed into the embedding layer, while the corresponding residual $\theta_r$ is discarded. This process ensures that information about the precise orientation is not entirely lost, as the residual component is preserved.

\noindent\textbf{Noisy Boxes Embedding.} Given the noisy continuous value $\left(\widetilde{l}_{3D}, \widetilde{h}_{3D}, \widetilde{w}_{3D}, \widetilde{d}\right)$, we apply sinusoidal positional encoding to map it into a high-dimensional vector. For the noisy discrete category $\widetilde{c}$ and orientation bin $\widetilde{\theta}_{bin}$, we utilize learnable embeddings. These encoded continuous features and discrete embeddings are then concatenated and subsequently fed into a three-layer Multi-Layer Perceptron (MLP) to obtain the noisy query vectors.

\noindent\textbf{3D Noisy Query Generator.} As described in Section 3.1 of the main manuscript, we introduce $Q_N = \{\{q_{N_{ij}}\}_{j=1}^C\}_{i=1}^G$ as noisy queries for the denoising process, where $q_{N_{ij}}\in \mathbb{R}^{K\times D}$ with $K$ being the number of objects in the input image. Prior to the addition of noise and feeding into the embedding layer, the ground truth object set is first repeated $C \times G$ times to obtain the exact number of queries needed.

\subsection{Mask Separated Self-Attention}
Recalling that Separated Self-Attention~\cite{GroupDETR} enables self-attention to aggregate information independently within each learnable group, we introduce an attention mask to further manage the aggregation between noisy and learnable queries.

\noindent\textbf{Mask Design.} Adhering to the rational constraint of preventing information leakage as in DN-DETR~\cite{DN-DETR}, we devise an attention mask that satisfies two conditions: (1) learnable queries must not aggregate noisy queries to ensure consistency between training and inference, given that noisy queries are discarded during inference; (2) noisy queries within each group must not attend to those of other groups.

As described in Equation 3 of the main manuscript, the new set of queries $Q=\{q_i\}_{i=q}^G$ is constructed before being fed into the Separated Self-Attention~\cite{GroupDETR}. Consequently, the attention map $M$ must have a shape of $S\times S$. The detailed design of $M$ can be expressed by the following formulation:

\begin{equation}
    M_{i,j} = 
     \left\{
     \begin{array}{l}
          1, \texttt{ if } j < K\times C \texttt{ and } \lfloor\frac{i}{C}\rfloor \neq \lfloor\frac{j}{C}\rfloor \\
          1, \texttt{ if } j < K\times C \texttt{ and } i \geq K\times C\\
          0, \texttt{ otherwise}
     \end{array}
     \right.
\end{equation}
where $M_{i,j}=1$ means the i-th query cannot access the j-th query and $M_{i,j}=0$ otherwise.

\noindent\textbf{Implementation Detail.} The pseudocode of Mask Separated Self-Attention is shown in Alg.~\ref{alg:mask_separated_self_attention}.

\begin{algorithm}
\caption{Pseudocode of Mask Separated Self-Attention}
\label{alg:mask_separated_self_attention}
\definecolor{codeblue}{rgb}{0.25,0.5,0.5}
\definecolor{codekw}{rgb}{0.85, 0.18, 0.50}
\lstset{
  backgroundcolor=\color{white},
  basicstyle=\fontsize{7.5pt}{7.5pt}\ttfamily\selectfont,
  columns=fullflexible,
  breaklines=true,
  captionpos=b,
  commentstyle=\fontsize{7.5pt}{7.5pt}\color{codeblue},
  keywordstyle=\fontsize{7.5pt}{7.5pt}\color{codekw},
}
\begin{lstlisting}[language=python]
# SA: Self-Attention in the decoder layer
# Q_L: learnable queries, with size (GxN, B, D)
# Q_N: noisy queries, with size (G,CxK,B,D)
# M: attention mask, with size (S,S)
# B: batch size

# Mask Separated Self-Attention
if training:
    # Construct a new set of queries Q
    Q_L = Q_L.split(N, dim = 0) # a list of G tensors with shape (N,B,D)
    Q_L = cat(Q_L,dim=1) # (N,BxG,D)
    Q_N = Q_N.split(1, dim = 0).squeeze(0) # a list of G tensors with shape (CxK,B,D)
    Q_N = cat(Q_N,dim=1) # (CxK,BxG,D)
    Q = cat((Q_N,Q_L),dim=0) # (CxK+N,BxG,D) or (S,BxG,D)

    # mask self-attention
    out = SA(Q,M) # (S,BxG,D)

    # Split the output back to learnable queries and noisy queries
    Q_N = cat(out[:-N].unsqueeze(0).split(B,dim=2),dim=0) # (G,CxK,B,D)
    Q_L = cat(out[-N:].split(B,dim=1),dim=0) # (GxN,B,D)
    
else:   
    # In inference phase Q_N = none, M = none
    Q = Q_L[:N]
    out = SA(Q)
\end{lstlisting}
\end{algorithm}
\subsection{Denoising Loss}
\textbf{Reconstruct Loss.} We uniformly applied the reconstruction loss to all bounding box properties, consistent with the formulation of $L_{det}$.

\noindent\textbf{Forward-Looking Distillation for Noisy Queries.} Unlike in the case of learnable queries, the requirement for Hungarian matching between the final decoder layer and ground truth to identify positive queries is obviated. For noisy queries, we instead implement a distillation loss for the queries at each layer, which are derived from the same ground truth.
\subsection{Hyperparameters}
We trained Mono3DV on a single NVIDIA 3090 GPU for 250 epochs. We used a batch size of 8 and an initial learning rate of $2 \times 10^{-4}$. Further training specifics are detailed in Table~\ref{tab:hyper}.
\begin{table}[h!]
    \centering
    \begin{tabular}{ll}
         \toprule
         \textbf{Item} & \textbf{Value} \\
         \midrule
         $\lambda_1$ & 1  \\
         $\lambda_2$ & 1 \\
         $\lambda_3$ & 0.5 \\
         $\lambda_C$ & 0.4 \\
         $\lambda_D$ & 0.2 \\
         $\epsilon$ & 1 \\
         $T$ & 85 \\
         $N$ & 50 \\
         $G$ & 11\\
         $C$ & 5 \\
         $\beta$ & 4\\
         weight decay & 1e-4\\
         scheduler & Step \\
         decay rate & 0.5 \\
         decay list & [85, 125, 165, 205] \\
         number of feature scales & 4 \\
         hidden dim & 256 \\
         feedforward dim & 256 \\
         dropout & 0.1 \\
         nheads & 8 \\
        number of encoder layers & 3 \\
        number of decoder layers & 3 \\
         encoder npoints  & 4 \\
         decoder points & 4\\
         \bottomrule
    \end{tabular}
    \caption{\textbf{Main hyperparameters of Mono3DV.}}
    \label{tab:hyper}
\end{table}
\section{Why Variational Query DeNoising?}
This section rigorously examines the underlying causes of the gradient vanishing problem as it pertains to the reconstruction loss's impact on the learnable queries and explains why the proposed Variational Query Denoising approach successfully overcomes this challenge. 

\noindent\textbf{Forward.} As described in Equation 3 of the main manuscript, the new set of queries $Q$ is constructed from the sets of learnable queries $Q_L$ and noisy queries $Q_N$. For convenience, we recall this notation here:
\begin{equation}
    \begin{aligned}
        Q &=\{q_i\}_{i=1}^G, \texttt{ where } q_i\in \mathbb{R}^{S\times D} \\
        Q_L &= \{q_{L_i}\}_{i=1}^G, \texttt{ where } q_{L_i}\in\mathbb{R}^{N \times D} \\
        Q_N &= \{\{q_{N_{ij}}\}_{j=1}^C\}_{i=1}^G, \texttt{ where } q_{N_{ij}}\in \mathbb{R}^{K\times D} \\
        q_i &= \texttt{concat}\left(q_{N_{i1}},q_{N_{i2}},...,q_{N_{iC}},q_{L_i}\right) \\
        & \hspace{6em} \texttt{ for } i\in \{1,2,...,G\}\\
    \end{aligned}
    \label{eq:recall_q_i}
\end{equation}

This newly constructed set $Q$ is then fed into the Mask Separated Self-Attention mechanism. To clearly distinguish the outputs of this mechanism from their inputs, we denote the output by prefacing the input notation with an $o$ at the base letter. The detailed calculations for the Mask Separated Self-Attention are as follows:
\begin{equation}
    \begin{aligned}
        o_{q_i} &= \texttt{concat}\left(o_{q_{N_{i1}}},o_{q_{N_{i2}}},...,o_{q_{N_{iC}}},o_{q_{L_i}}\right) \\
        o_{N_{ijk}} &= \underbrace{\sum_{m=1}^{K} A\left(u,v\right)q_{N_{ijm}}}_{\texttt{noisy queries aggregation}} \\ 
        &+ \underbrace{\sum_{m=1}^{N}A\left(u, w\right)q_{L_{im}}}_{\texttt{learnable queries aggregation}} \\ & \hspace{2em} \texttt{for } j\in \{1,2,...,C\},k\in \{1,2,...,K\} \\
        o_{q_{L_{im}}} &= \sum_{n=1}^N A\left(w, x\right)q_{L_{in}} \\
        & \hspace{3em} \texttt{for } m \in \{1,2,...,N\} \\
        & \texttt{with } u = (j-1)\times K + k,\\
        & \hspace{3.2em} v = (j-1)\times K + m,\\
        & \hspace{3.2em} w = (C-1)\times K + m,\\
        & \hspace{3.2em} x = (C-1)\times K + n\\
    \end{aligned}
    \label{eq:forward_msa}
\end{equation}
where $A \in [0,1]^{S \times S}$ denotes the attention map, $A(u,v)$ refers to the attention score at row $u$ and column $v$.

These outputs are subsequently fed into the remaining part of the decoder and the detection head to compute the loss function. It is important to note that only the outputs of the noisy queries $o_{N_{ijk}}$, are utilized for calculating the reconstruction loss.

\textbf{Backward.} Based on the preceding discussion, the gradient of the reconstruction loss $L_{res}$ with respect to the learnable queries $Q_L$ can be expressed as:
\begin{equation}
    \begin{aligned}
        \frac{\partial L_{res}}{\partial q_{L_{im}}} &= \sum_{j=1}^C\sum_{k=1}^K\frac{\partial L_{res}}{\partial o_{N_{ijk}}} \frac{\partial o_{N_{ijk}}}{\partial q_{L_{im}}} \\
        &= \sum_{j=1}^C\sum_{k=1}^K\frac{\partial L_{res}}{\partial o_{N_{ijk}}} A\left(u,w\right) \\
        & \hspace{1em} \texttt{for } i\in \{1,2,...,G\}, m\in \{1,2,...,N\}
    \end{aligned}
    \label{eq:backward_msa}
\end{equation}
As illustrated in Figure 3(a) of the main manuscript, in conventional denoising approaches, the attention scores between noisy queries and learnable queries, specifically $A\left(u,w\right)$ , tend to approach zero. This consequently leads to:
\begin{equation}
    \frac{\partial L_{res}}{\partial q_{L_{im}}}\approx 0
\end{equation}
which causes a gradient vanishing problem from the reconstruction loss to the learnable queries.

To address this vanishing gradient issue, a straightforward solution is to increase the attention score $A\left(u,w\right)$. As demonstrated in Figure 3(b) of the main manuscript, the Variational Query DeNoising approach indeed yields larger attention scores between noisy queries and learnable queries, thereby enabling the model to effectively overcome the gradient vanishing problem.
\section{Experiments on Waymo Open Dataset}
The Waymo~\cite{waymo} dataset evaluates object detection by classifying objects as Level\_1 and Level\_2, which are determined by the number of LiDAR points
within their 3D bounding boxes. The experiments is conducted across three distance ranges: $[0, 30)$. $[30, 50)$, and $[50,\infty)$ meters. Performance on the Waymo dataset is assessed by average precision $AP_{3D}$ and average precision
weighted by heading $APH_{3D}$.

We follow the DEVIANT~\cite{DEVIANT} split to generate 52,386
training and 39,848 validation images by sampling every
third frame. For fairness, we mainly compare with methods
using the same split in \cref{tab:waymo_experiment}. Our method achieves stateof-the-art performance without extra data across all ranges,
particularly for distant objects. These results further validate the effectiveness and generalizability of Mono3DV.
\begin{table*}[!ht]
\centering
\caption{Results on the Waymo val set for the vehicle category. Compared with methods without extra data, we \textbf{bold} the best results and {\ul underline} the second-best results.}
\resizebox{0.90\textwidth}{!}{%
\begin{tabular}{c|l|c|cccc|cccc}
\hline
\multirow{2}{*}{Difficulty} & \multirow{2}{*}{Methods} & \multirow{2}{*}{Extra} & \multicolumn{4}{c|}{$AP_{3D}$} & \multicolumn{4}{c}{$APH_{3D}$} \\ 
\cline{4-11} 
 &  &  & All & 0-30 & 30-50 & 50-$\infty$ & All & 0-30 & 30-50 & 50-$\infty$ \\ 
\hline
\multirow{8}{*}{{Level\_1 \newline (IoU=0.7)}} 
 & PatchNet \cite{patchnet} in \cite{pct} & Depth & 0.39 & 1.67 & 0.13 & 0.03 & 0.39 & 1.63 & 0.12 & 0.03 \\
 & PCT \cite{pct} & Depth & 0.89 & 3.18 & 0.27 & 0.07 & 0.88 & 3.15 & 0.27 & 0.07 \\
 & M3D-RPN \cite{m3d-rpn} in \cite{caddn} & None & 0.35 & 1.12 & 0.18 & 0.02 & 0.34 & 1.10 & 0.18 & 0.02 \\
 & GUPNet \cite{GUPNet} in \cite{DEVIANT} & None & 2.28 & 6.15 & 0.81 & 0.03 & 2.27 & 6.11 & 0.80 & 0.03 \\
 & DEVIANT \cite{DEVIANT} & None & 2.69 & 6.95 & 0.99 & 0.02 & 2.67 & 6.90 & 0.98 & 0.02 \\
 & MonoUNI \cite{MonoUNI} & None & 3.20 & 8.61 & 0.87 & 0.13 & 3.16 & 8.50 & 0.86 & 0.12 \\
 & MonoDGP \cite{MonoDGP} & None & {\ul 4.28} & {\ul10.24} & {\ul1.15} & {\ul0.16} & {\ul4.23} & {\ul10.10} & {\ul1.14} & {\ul0.16} \\ 
 & \textbf{Mono3DV (Ours)} & None & \textbf{4.84} & \textbf{11.02} & \textbf{1.28} & \textbf{0.19} & \textbf{4.68} & \textbf{10.78} & \textbf{1.23} & \textbf{0.18} \\
\hline
\multirow{8}{*}{{Level\_2\newline (IoU=0.7)}} 
 & PatchNet \cite{patchnet} in \cite{pct} & Depth & 0.38 & 1.67 & 0.13 & 0.03 & 0.36 & 1.63 & 0.11 & 0.03 \\
 & PCT \cite{pct} & Depth & 0.66 & 3.18 & 0.27 & 0.07 & 0.66 & 3.15 & 0.26 & 0.07 \\
 & M3D-RPN \cite{m3d-rpn} in \cite{caddn} & None & 0.35 & 1.12 & 0.18 & 0.02 & 0.33 & 1.10 & 0.17 & 0.02 \\
 & GUPNet \cite{GUPNet} in \cite{DEVIANT} & None & 2.14 & 6.13 & 0.78 & 0.02 & 2.12 & 6.08 & 0.77 & 0.02 \\
 & DEVIANT \cite{DEVIANT} & None & 2.52 & 6.93 & 0.95 & 0.02 & 2.50 & 6.87 & 0.94 & 0.02 \\
 & MonoUNI \cite{MonoUNI} & None & 3.04 & 8.59 & 0.85 & 0.12 & 3.00 & 8.48 & 0.84 & 0.12 \\
 & MonoDGP \cite{MonoDGP} & None & {\ul4.00} & {\ul10.20} & {\ul1.13} & {\ul0.15} & {\ul3.96} & {\ul10.08} & {\ul1.12} & {\ul0.15} \\ 
 & \textbf{Mono3DV (Ours)} & None & \textbf{4.55} & \textbf{10.89} & \textbf{1.31} & \textbf{0.17} & \textbf{4.46} & \textbf{10.68} & \textbf{1.20} & \textbf{0.17} \\
\hline
\multirow{8}{*}{{Level\_1\newline (IoU=0.5)}} 
 & PatchNet \cite{patchnet} in \cite{pct} & Depth & 2.92 & 10.03 & 1.09 & 0.23 & 2.74 & 9.75 & 0.96 & 0.18 \\
 & PCT \cite{pct} & Depth & 4.20 & 14.70 & 1.78 & 0.39 & 4.15 & 14.54 & 1.75 & 0.39 \\
 & M3D-RPN \cite{m3d-rpn} in \cite{caddn} & None & 3.79 & 11.14 & 2.16 & 0.26 & 3.63 & 10.70 & 2.09 & 0.21 \\
 & GUPNet \cite{GUPNet} in \cite{DEVIANT} & None & 10.02 & 24.78 & 4.84 & 0.22 & 9.94 & 24.59 & 4.78 & 0.22 \\
 & DEVIANT \cite{DEVIANT} & None & 10.98 & 26.85 & 5.13 & 0.18 & 10.89 & 26.64 & 5.08 & 0.18 \\
 & MonoUNI \cite{MonoUNI} & None & 10.98 & 26.63 & 4.04 & 0.57 & 10.73 & 26.30 & 3.98 & 0.55 \\
 & MonoDGP \cite{MonoDGP} & None & {\ul12.36} & {\ul31.12} & {\ul5.78} & {\ul1.24} & {\ul12.18} & {\ul30.68} & {\ul5.71} & {\ul1.22} \\ 
 & \textbf{Mono3DV (Ours)} & None & \textbf{13.53} & \textbf{34.72} & \textbf{6.11} & \textbf{1.59} & \textbf{13.28} & \textbf{33.81} & \textbf{6.19} & \textbf{1.57} \\
\hline
\multirow{8}{*}{{Level\_2\newline (IoU=0.5)}} 
 & PatchNet \cite{patchnet} in \cite{pct} & Depth & 2.42 & 10.01 & 1.07 & 0.22 & 2.28 & 9.73 & 0.97 & 0.16 \\
 & PCT \cite{pct} & Depth & 4.03 & 14.67 & 1.74 & 0.36 & 4.15 & 14.51 & 1.71 & 0.35 \\
 & M3D-RPN \cite{m3d-rpn} in \cite{caddn} & None & 3.61 & 11.12 & 2.12 & 0.24 & 3.46 & 10.67 & 2.04 & 0.20 \\
 & GUPNet \cite{GUPNet} in \cite{DEVIANT} & None & 9.39 & 24.69 & 4.67 & 0.19 & 9.31 & 24.50 & 4.62 & 0.19 \\
 & DEVIANT \cite{DEVIANT} & None & 10.29 & 26.75 & 4.95 & 0.16 & 10.20 & 26.54 & 4.90 & 0.16 \\
 & MonoUNI \cite{MonoUNI} & None & 10.38 & 26.57 & 3.95 & 0.53 & 10.24 & 26.24 & 3.89 & 0.51\\
 & MonoDGP \cite{MonoDGP} & None & {\ul11.71} & {\ul31.02} & {\ul5.61} & {\ul1.17} & {\ul11.56} & {\ul30.58} & {\ul5.54} & {\ul1.15} \\ 
  & \textbf{Mono3DV (Ours)} & None & \textbf{12.92} & \textbf{34.65} & \textbf{5.93} & \textbf{1.48} & \textbf{12.76} & \textbf{34.08} & \textbf{5.87} & \textbf{1.46} \\
\hline 
\end{tabular} 
}
\label{tab:waymo_experiment}
\end{table*}

\section{Multi-view 3D Object Detection Experiments}
The nuScenes~\cite{nuscenes} dataset is composed of 1000 video sequences, divided into 700 for training, 150 for validation, and 150 for testing. Each sequence is approximately 20 seconds long with annotations provided at 0.5-second intervals. Performance is evaluated using the mean Average Precision (mAP) and five true positive metrics: ATE, ASE, AOE, AVE, and AAE, which respectively measure errors in translation, scale, orientation, velocity, and attribute prediction. These metrics are combined to form the comprehensive nuScenes Detection Score (NDS), providing an overall evaluation of performance.

We build upon the MonoDETR baseline~\cite{MonoDETR} by conducting a multi-view plug-and-play study. This involves integrating Variational Query DeNoising and Forward-Looking Distillation onto two DETR-based multi-view networks: RayDN~\cite{RayDN} and OPEN~\cite{OPEN}. Since these DETR-based multi-view baselines already incorporate a denoising technique, our modification was simplified to adapting the query embedding as a Variational Autoencoder (VAE) and adding a self-distillation loss. For RayDN~\cite{RayDN}, the approach delivered +0.8\% NDS and +0.9\% mAP by optimizing learnable query denoising and iterative refinement strategies. Applied to OPEN~\cite{OPEN}, it achieved +1.2\% NDS and +1.1\% mAP, successfully mitigating conventional denoising's gradient vanishing issues while boosting iterative refinement. These multi-view 3D object detection results confirm the approach's effectiveness and generalizability.
\begin{table*}[ht]    
  \centering
  \caption{\textbf{Comparison on the nuScenes validation set.} $\dagger$ Indicates methods that benefit from perspective-view pre-training.}
\begin{tabular}{l|cc|cccccc}
\toprule
\textbf{Methods} & \textbf{mAP}$\uparrow$  &\textbf{NDS}$\uparrow$ & \textbf{mATE}$\downarrow$& \textbf{mASE}$\downarrow$& \textbf{mAOE}$\downarrow$& \textbf{mAVE}$\downarrow$& \textbf{mAAE}$\downarrow$ \\
\midrule
PETRv2~\cite{liu2023petrv2} &  34.9    & 45.6    & 0.700  & 0.275& 0.580 & 0.437& 0.187\\
BEVDepth~\cite{li2023bevdepth}    &   35.1    & 47.5    & 0.629& 0.267& 0.479& 0.428& 0.198\\
% BEVPoolv2~\cite{huang2022bevpoolv2}   &   40.6    & 52.6    & 0.572& 0.275& 0.463& 0.275& 0.188\\
SOLOFusion~\cite{park2022time}  &   42.7    & 53.4    & 0.567    & 0.274& 0.511& 0.252& 0.181    \\
SparseBEV~\cite{liu2023sparsebev}$\dagger$   &  44.8    & 55.8    & 0.581& 0.271& 0.373   & 0.247    & 0.190 \\
StreamPETR~\cite{Wang_2023_ICCV}$\dagger$  &   45.0& 55.0& 0.613& 0.267& 0.413& 0.265& 0.198\\ \midrule
RayDN$\dagger$~\cite{RayDN}  &  46.9 & 56.3 & 0.579 & 0.264 & 0.433 & 0.256 & 0.187 \\
\hspace{-1pt} + \textit{Variational DeNoising} & \textbf{47.8} & \textbf{57.1} & 0.571 & 0.262 & 0.391 & 0.273 & 0.180 \\ \midrule
OPEN$\dagger$~\cite{OPEN} & 46.5 & 56.4 & 0.573 & 0.275 & 0.413 & 0.235 & 0.193 \\
\hspace{-1pt} + \textit{Variational DeNoising} & \textbf{47.6} & \textbf{57.6} & 0.564 & 0.271 & 0.384 & 0.206 & 0.197 \\
\bottomrule
\end{tabular}
\label{tab:nuscence_result}
% \vspace{-4pt}
\end{table*}
\section{Additional Ablation Study}
\textbf{Hyperparameter choosing for 3D Weight Scheduler.} For 3D Weight Scheduler, we set distinct value for the weight threshold $\epsilon$ and trigger epoch $T$. Optimal performance was achieved by setting the integration parameters to $\epsilon=1$ and $T=85$. This specific parameter configuration effectively integrates the 3D attribute into the Bipartite Matching algorithm at the most opportune time and with the appropriate weighting, resulting in the highest reported performance.
\begin{table}[h!]
    \centering
    \caption{\textbf{The design of step scheduler in 3D Weight Scheduler.} $\epsilon$ denotes the weight threshold and $T$ is the trigger epoch.}
    \begin{tabular}{cc|ccc}
         \toprule
         \multicolumn{2}{c|}{Ablation} & \multicolumn{3}{c}{Val, $AP_{3D|R40}$} \\
         $\epsilon$ & $T$ & Easy & Mod. & Hard \\ \midrule \midrule
         0.1 & 165 &  29.52 & 21.13 & 17.92 \\
         1 & 165  & 29.86 & 21.45 & 18.11\\
         0.1 & 125 & 29.67 & 21.36 & 17.98 \\
         1 & 125 & 29.95 & 21.58 & 18.14 \\
         0.1 & 85 & 29.84 & 21.43 & 18.07 \\
         1 & 85 &\textbf{30.09} & \textbf{21.63} & \textbf{18.27} \\
         \bottomrule
    \end{tabular}
    \label{tab:3d_weight_scheduler_step}
\end{table}

\noindent\textbf{Number of group denoising.} We explore the influence of the number of denoising groups on model performance in Tab.~\ref{tab:denoising_group_ablation}. As detailed in Section 3.1 of the main manuscript, we establish $C$ denoising groups for each of $K$ learnable query groups. Our findings reveal that increasing the number of denoising groups generally enhances performance. However, this improvement diminishes gradually as the number of groups grows. Consequently, we employ $C=5$ as our default setting in our experiments for appropriate with the hardware ability.
\begin{table}[h!]
    \centering
    \caption{\textbf{Comparison on the effect of number denoising group} to the model. $C=1$ denotes using only one denoising group for each learnable query group. $C=5$ denotes duplicating the ground truth 5 times and creating 5 denoising groups for each learnable query group.}
    \begin{tabular}{c|ccc}
         \toprule
         \multirow{2}{*}{\# of denoising groups} & \multicolumn{3}{c}{Val, $AP_{3D|R40}$} \\
          & Easy & Mod. & Hard \\ \midrule \midrule
         C = 1 & 30.28 & 22.18 & 18.70\\
         C = 5 & \textbf{30.46} & \textbf{22.78} & \textbf{19.49} \\
         \bottomrule
    \end{tabular}
    \label{tab:denoising_group_ablation}
\end{table}

\noindent\textbf{Design of Noisy Boxes Embedding.} We studied how the number of layers in the noisy boxes embedding within our 3D Noisy Query Generator affects performance. As detailed in ~\cref{tab:mlp_noisy_embed}, model performance steadily improved as we increased the MLP layers from one to three. However, a slight decrease in performance was observed with a 4-layer MLP, which we attribute to the limited sample size of the KITTI dataset.
\begin{table}[h!]
    \centering
    \caption{\textbf{The design of Noisy Boxes Embedding.} After applying sinusoidal positional encoding to continuous noisy values and a learnable embedding for discrete noisy values, we initially constructed the noisy query using a single linear layer. We later explored alternative configurations, including a two-layer MLP and progressively deeper variants.} 
    \begin{tabular}{c|ccc}
         \toprule
         \multirow{2}{*}{\# of MLP Layers} & \multicolumn{3}{c}{Val, $AP_{3D|R40}$} \\
          & Easy & Mod. & Hard \\ \midrule \midrule
         1 &  30.32 & 22.27 & 18.93 \\
         2 & 30.41 & 22.53 & 19.21\\
         3 & \textbf{30.46} & \textbf{22.78} & \textbf{19.49} \\
         4 & 30.36 & 22.45 & 19.01 \\
         \bottomrule
    \end{tabular}
    \label{tab:mlp_noisy_embed}
\end{table}

\noindent\textbf{Forward-Looking Distillation.} We analyze the efficacy of each design component of Forward-Looking distillation in ~\cref{tab:self_distillation}. Firstly, incorporating $\mathbf{IoU}_{3D}$ weighting mechanism to ensure quality distillation knowledge from the good query, we can see that the model achieves marked performance. Moreover, using an MLP to refine the early decoder query also improves the detection quality.
\begin{table}[h!]
    \centering
        \caption{\textbf{The design of Forward-Looking distillation.} w/o denotes directly aligned early query with the last one.`$\mathbf{IoU}_{3D}$' denotes using $\mathbf{IoU}_{3D}$ for weighting between queries. `MLP' denotes using a two-layer MLP to refine the student query.} 
    \begin{tabular}{c|ccc}
         \toprule
         \multirow{2}{*}{Ablation} & \multicolumn{3}{c}{Val, $AP_{3D|R40}$} \\
          & Easy & Mod. & Hard \\ \midrule \midrule
         w/o &  29.25 & 21.12 & 17.65 \\
         $\mathbf{IoU}_{3D}$ & 29.71 & 21.32 & 18.05\\
         $\mathbf{IoU}_{3D}$ + MLP & \textbf{30.05} & \textbf{21.54} & \textbf{18.31} \\
         \bottomrule
    \end{tabular}
    \label{tab:self_distillation}
\end{table}

\noindent\textbf{Design of Refinement MLP.} We explore the impact of a number of layers of the refinement MLPs in Forward-Looking Distillation on model performance. As shown in ~\cref{tab:refine_mlp_distill}, our default 2-layer MLP consistently delivered the best overall results. Deeper MLP configurations, however, performed poorly, likely due to the limited sample size of the KITTI dataset.
\begin{table}[h!]
    \centering
    \caption{\textbf{The design of Refinement MLP in Forward-Looking Distillation.} Initially, we directly align the early query with the last one, denoted as "w/o" for without explicit refinement. Subsequently, the performance was evaluated across various MLP architectures, including a single linear layer, our default two-layer configuration, and progressively deeper MLP structures, to understand their impact on the distillation process.} 
    \begin{tabular}{c|ccc}
         \toprule
         \multirow{2}{*}{\# of MLP Layers} & \multicolumn{3}{c}{Val, $AP_{3D|R40}$} \\
          & Easy & Mod. & Hard \\ \midrule \midrule
         w/o & 29.71 & 21.32 & 18.05 \\
         1 &  29.84 & 21.46 & 18.13 \\
         2 & \textbf{30.05} & \textbf{21.54} & \textbf{18.31}\\
         3 & 29.31 & 21.24 & 17.94 \\
         \bottomrule
    \end{tabular}
    \label{tab:refine_mlp_distill}
\end{table}
\section{Additional Qualitative Results}
\begin{figure*}[h!]
    \centering
    \includegraphics[width=\textwidth]{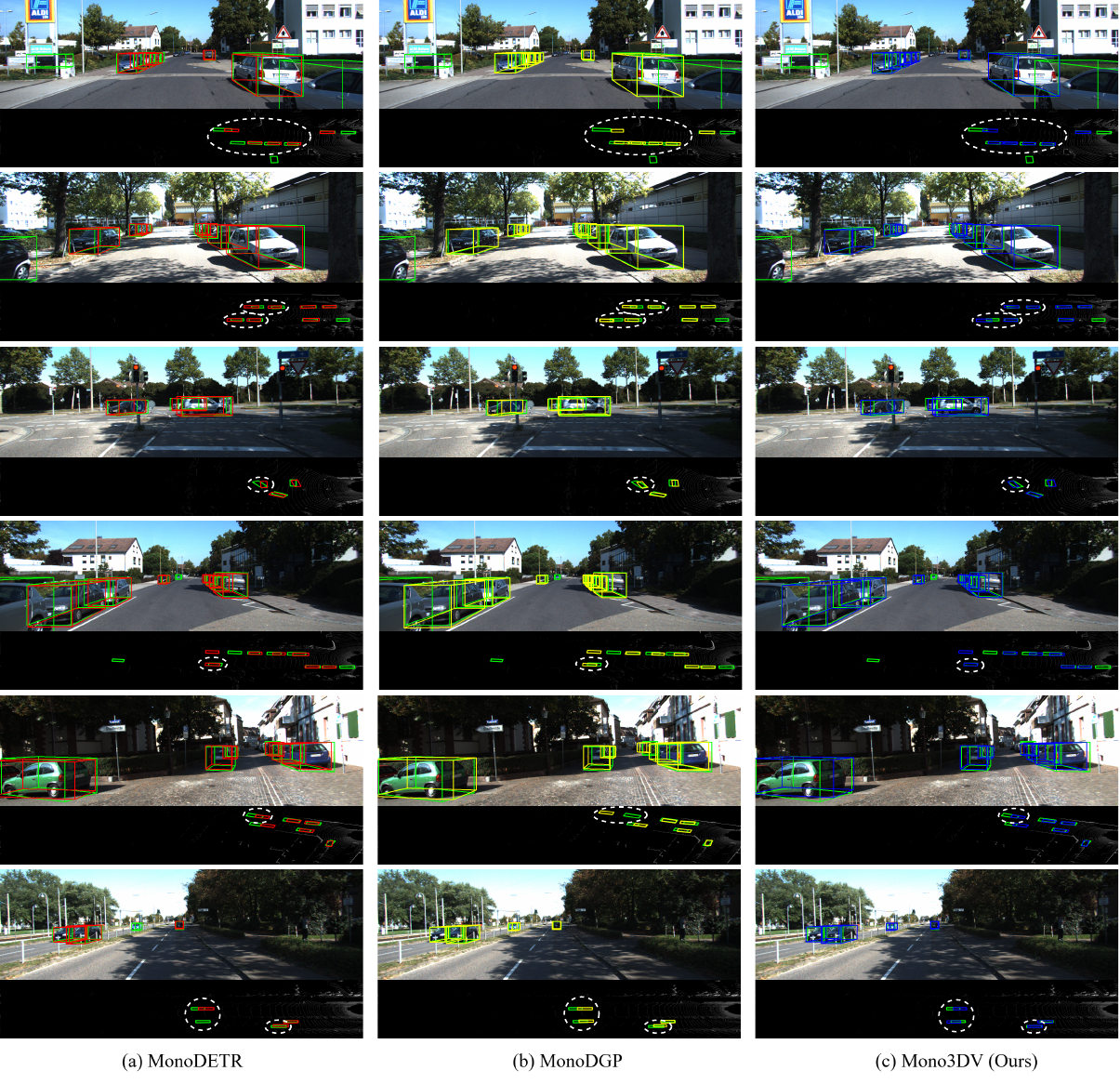}
    \caption{\textbf{Qualitative results on KITTI val set.} (a) MonoDETR~\cite{MonoDETR}. (b) MonoDGP~\cite{MonoDGP}. (c) Mono3DV (Ours). For each image set, the top row presents the camera-view visualization, while the bottom row offers the corresponding bird’s-eye view. Ground-truth bounding boxes are rendered in \textcolor{green}{green}, and predictions are shown in order: \textcolor{red}{red}, \textcolor{yellow}{yellow}, \textcolor{blue}{blue}. We also circle some objects to highlight the difference between other state of the art and our method.}
\label{fig:additional_qualitative_results}
\end{figure*}

To provide a more intuitive comparison between our
method and other state-of-the art methods, we visualize some 3D detection results from both the camera view and the bird’s-eye view on the KITTI validation set. As shown in \cref{fig:additional_qualitative_results}, our
method demonstrates superior performance on distant and occluded objects, thereby demonstrating its superior robustness under challenging samples.
\end{document}